\documentclass{article}

\usepackage{microtype}
\usepackage{graphicx}
\usepackage{subfigure}
\usepackage{booktabs} 
\usepackage{arydshln}
\usepackage{hyperref}
\usepackage[table,xcdraw]{xcolor}
\usepackage{inconsolata}
\usepackage{xcolor}         
\usepackage{graphicx}
\usepackage{amsmath}
\usepackage{multirow}
\usepackage{subcaption}
\usepackage{makecell}
\usepackage{tabularx}
\usepackage{wrapfig}
\usepackage{balance}

\usepackage{hyperref}       
\usepackage{url}            
\usepackage{booktabs}       
\usepackage{amsfonts}       
\usepackage{nicefrac}       
\usepackage{microtype}      

\definecolor{myblue}{rgb}{0.8235, 0.9333, 0.9412}

\usepackage{amsmath}
\usepackage{amssymb}
\usepackage{mathtools}
\usepackage{amsthm}
\usepackage{listings}
\usepackage[capitalize,noabbrev]{cleveref}
\usepackage{soul, color, xcolor}

\usepackage[accepted]{icml2025}

\theoremstyle{plain}

\theoremstyle{definition}

\theoremstyle{remark}

\usepackage[textsize=tiny]{todonotes}

\icmltitlerunning{MergeQuant: Accurate 4-bit Static Quantization of Large Language Models by Channel-wise Calibration}

\begin{document}

\twocolumn[
\icmltitle{MergeQuant: Accurate 4-bit Static Quantization of Large Language Models by Channel-wise Calibration}



\icmlsetsymbol{equal}{*}

\begin{icmlauthorlist}
\icmlauthor{Jinguang Wang}{yyy,comp}
\icmlauthor{Jingyu Wang}{yyy,comp}
\icmlauthor{Haifeng Sun}{yyy,equal}
\icmlauthor{Tingting Yang}{comp}
\icmlauthor{Zirui Zhuang}{yyy}
\icmlauthor{Wanyi Ning}{yyy}
\icmlauthor{Yuexi Yin}{yyy,comp}
\icmlauthor{Qi Qi}{yyy}
\icmlauthor{ Jianxin Liao}{yyy}

\end{icmlauthorlist}

\icmlaffiliation{yyy}{State Key Laboratory of Networking and Switching Technology, Beijing University of Posts and Telecommunications, Beijing, China}
\icmlaffiliation{comp}{PengCheng Laboratory, Shenzhen, China}


\vskip 0.3in
]



\footnote{
    State Key Laboratory of Networking and Switching Technology, Beijing University of Posts and Telecommunications, Beijing, China; 
    \textsuperscript{2}PengCheng Laboratory, Shenzhen, China. Correspondence to: Haifeng Sun: hfsun@bupt.edu.cn.
}


\begin{abstract}
Quantization has been widely used to compress and accelerate inference of large language models (LLMs). 
Existing methods focus on exploring the per-token dynamic calibration to ensure both inference acceleration and model accuracy under 4-bit quantization. However, in autoregressive generation inference of long sequences, the overhead of repeated dynamic quantization and dequantization steps becomes  considerably expensive. In this work, we propose MergeQuant, an accurate and efficient per-channel static quantization framework. MergeQuant integrates the per-channel quantization steps with the corresponding scalings and linear mappings through a Quantization Step Migration (QSM) method, thereby eliminating the quantization overheads before and after matrix multiplication.
Furthermore, in view of the significant differences between the different channel ranges, we propose dimensional reconstruction and adaptive clipping to address the non-uniformity of quantization scale factors and redistribute the channel variations to the subsequent modules to balance the parameter distribution under QSM. Within the static quantization setting of W4A4, MergeQuant reduces the accuracy gap on zero-shot tasks compared to FP16 baseline to 1.3 points on Llama-2-70B model. On Llama-2-7B model, MergeQuant achieves  up to 1.77× speedup in decoding, and up to 2.06× speedup in end-to-end  compared to FP16 baseline.  
\end{abstract}

\section{Introduction}
\label{submission}
Large Language Models (LLMs) have garnered widespread attention due to their remarkable advancements and exceptional performance in complex language generation tasks \cite{vaswani2017attention, touvron2023llama, radford2019language, NEURIPS2022_77c6ccac, zhang2022opt, brown2020language}. Nevertheless, the substantial computational and memory demands of these models during autoregressive inference pose significant challenges to their broad application, especially on edge devices with weak computing and storage resources \citep{liu2023scissorhands, pope2022efficiently}. An effective solution is to reduce the deployment cost through model quantization \cite{gholami2021survey, wang2019haq, tao-etal-2022-compression, zhang-etal-2020-ternarybert}, which reduces model size and computational demands by converting weights and activations into lower-precision integer representations \cite{bai-etal-2021-binarybert, pmlr-v139-kim21d}. For instance, quantizing weights and activations to 4-bit integers can significantly reduce the memory footprint, making it feasible to deploy a Llama-2-70B model with just 40 GB of memory.

Model quantization can be categorized into dynamic quantization and static quantization according to the quantization types of activations. Dynamic quantization calculates quantization parameters during model inference in real-time \cite{gholami2021survey,wang2022fine}. In scenarios involving  small-batch autoregressive inference with long sequences, the repeated quantization and dequantization steps of dynamic quantization introduce additional computational and data movement for each input token, making such overhead  particularly expensive \cite{NEURIPS2022_adf7fa39}. In contrast, static quantization  pre-computes and stores all necessary quantization parameters prior to inference, thereby eliminates the additional computational and data movement overheads associated with  quantization and dequantization steps \cite{pmlr-v202-xiao23c, wang2024outliertune}.

Even though  the static quantization effectively avoids the additional inference overhead associated with the dynamic quantization \cite{pmlr-v202-xiao23c}, the low-bit static quantization of LLMs remains challenging. A major reason is that the static quantization requires calibration of quantization parameters along the tensor dimension, which is difficult for LLMs with structured outliers \cite{dettmers2022llmint8, NEURIPS2022_adf7fa39, liu2023qllm}. Current  dynamic quantization methods \cite{ashkboos2024quarot, shao2023omniquant} employ pre-quantization transformations to suppress outliers and effectively mitigate quantization-induced performance losses by calibrating quantization parameters online along the token dimension. Unfortunately, these methods have difficulty maintaining the model performance under per-tensor calibration, making them challenging to extend to static quantization. To the best of our knowledge, only a few methods \cite{pmlr-v202-xiao23c, wei-etal-2023-outlier} have applied 6/8-bit static quantization on LLMs, but they fail to preserve the model performance under 4-bit static quantization. It therefore remains an open question whether 4-bit static quantization for LLMs is feasible.

In this paper, we introduce MergeQuant, an accurate and efficient per-channel static quantization framework designed for LLMs, enabling  4-bit quantization without significant performance losses. To address the challenge posed by structured outliers, we perform offline statistical analysis of activations along the channel dimension and assign independent quantization scaling factors to each activation channel. Specifically, we propose a Quantization Step Migration (QSM) method that integrates the per-channel quantization steps with corresponding scaling and linear mapping. This integration facilitates the alignment of matrix multiplication (MM) with the integer acceleration kernels, effectively reducing the per-channel quantization overhead before and after MM. To balance the parameter distribution in the linear mapping under QSM, we employ dimension reconstruction and adaptive clipping to alleviate the non-uniformity of activation quantization scaling factors and redistribute the channel variations to the subsequent modules. Building on this foundation, we further enhance the framework by introducing a small set of learnable quantization compensation parameters within the linear mapping, which compensate for the accuracy losses caused by per-channel quantization and clipping while ensuring quantization efficiency. 

To summarize, our main contributions are:
\begin{itemize}
\item[1)] We propose a novel per-channel static quantization framework that maintains the performance of 4-bit quantized models while achieving  comparable speed to per-tensor static quantization. This is made possible by a simple quantization step migration method, which facilitates the alignment of MM and integer acceleration kernels in per-channel activation quantization and avoids the necessary quantization step overheads.
\item[2)] To address the non-uniformity of activation quantization scaling factors and balance the parameter distribution within the linear mapping under QSM, we introduce dimension reconstruction and adaptive clipping. These methods mitigate the differences in quantization factors across channels, enabling the migration module to be equally suitable for quantization.
\item[3)] By incorporating learnable quantization compensation parameters, MergeQuant narrows the accuracy gap between the quantized Llama-2-70B model and FP16 baseline to just 1.3 points on zero-shot tasks. On Llama-2-7B model, MergeQuant achieves up to 2.72x speedup in pre-filling, up to 1.77x speedup in decoding, and up to 2.06x speedup in end-to-end.
\end{itemize}

\section{Preliminaries}
Quantization maps floating-point values to a fixed integer range, thereby reducing memory usage and improving computational efficiency. We investigate 4-bit static quantization to simplify model deployment and accelerate inference. The quantization process is as follows:
\begin{equation}\label{a}
\begin{aligned}
X^{Int} = \left\lceil\frac{X^{FP}}{s}\right\rfloor,\  s = \frac{max(|X^{FP}|)}{2^{b-1}-1},
\end{aligned}
\end{equation}
where $X^{Int}$ and $X^{FP}$ are integer and floating-point tensors, $\left\lceil\right\rfloor$ is the rounding function,  $s$ denotes the quantization scaling factors, and $b$ is the bits for integer mapping.  

Consider a linear layer in Transformer \cite{vaswani2017attention} $Y = XW$, where the weight $W\in \mathbb{R}^{n\times j}$ and activation $X\in \mathbb{R}^{i\times n}$ are both floating-point tensors. Based on \eqref{a}, we can accelerate the inference of linear layer through integer acceleration kernels (e.g., INT4 GEMM), as follows:
\begin{equation}\label{b}
\begin{aligned}
\hat{Y} = \left(\left\lceil\frac{X^{FP}}{s_X}\right\rfloor \left\lceil\frac{W^{FP}}{s_W}\right\rfloor\right)s_Xs_W,
\end{aligned}
\end{equation}
where the integer tensors are utilized for MM to accelerate the model inference, and then the result is de-quantized to ensure the equivalence with the FP network.
 
Determining the quantization scale  is crucial for accurate data representation \cite{wei2022outlier}. Static quantization uses a set of calibration samples to determine a fixed scale, ensuring minimal inference overhead. In contrast, dynamic quantization offers a more flexible alternative by adapting the scale in real-time to accommodate changing data conditions, thereby improving accuracy \cite{pmlr-v202-xiao23c}. Furthermore, the granularity of quantization calibration greatly impacts  the accuracy of data representation. Per-tensor quantization simplifies the quantization process by applying a single uniform scale to all elements within a tensor. To further ensure model performance, using finer-grained methods like per-token or per-channel quantization can mitigate quantization losses, though this adds complexity to the quantization process \cite{dettmers2022llmint8}.

\section{Difficulty of 4-bit Static Quantization}\label{sec3} 

 \begin{figure}[tbp]
    \centering
    \includegraphics[width=0.52\textwidth, trim=10 0 0 20, clip]{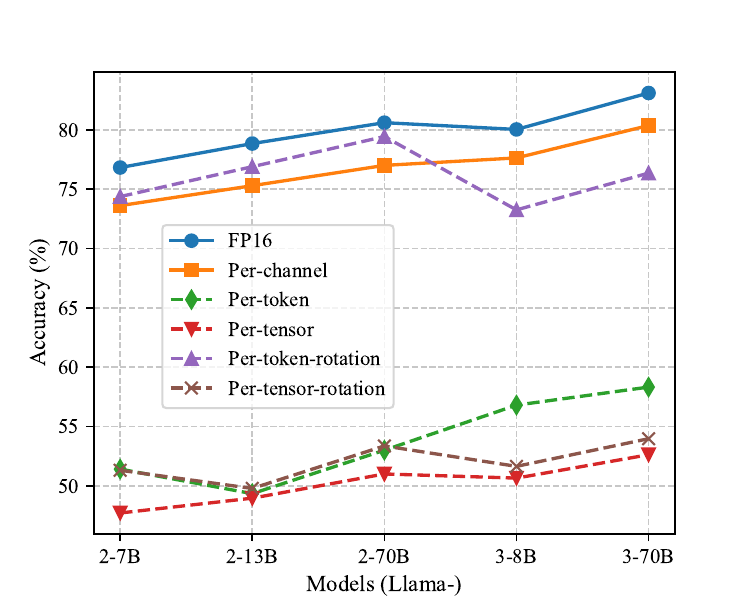}
    \caption{We evaluate the accuracy of three different calibrations and show the results when combined with rotation-based method in Llama models, measured on PIQA.}
    \label{fig1}\vspace{-5pt}
\end{figure} 

LLMs exhibit a distinct structured pattern of activation outliers, where the large activation outliers are concentrated in only a few channels \cite{dettmers2022llmint8, NEURIPS2022_adf7fa39}. These outliers dominate the numerical range of the entire data space during  the per-tensor and per-token calibrations and are the main cause of quantization round-off errors, as described in Appendix \ref{SecD}. In Fig. \ref{fig1}, we visualize the experimental results of different calibration methods. Under 4-bit symmetric quantization, the per-tensor and per-token calibrations lead to significant performance losses, while only the per-channel calibration  \cite{yuan2023rptq,wu2023understanding} maintains the model accuracy. This effectiveness is mainly due to the per-channel calibration method's ability to isolate the outlier channels, thus avoiding the outlier-induced adverse rounding of other normal values in the data space \cite{pmlr-v202-xiao23c}. Notably, since the per-tensor quantization represents the entire data space using only a single scale, even the state-of-the-art rotation-based methods \cite{ashkboos2024quarot, liu2024spinquant}, designed to smooth outliers,  still cannot guarantee the model accuracy under the per-tensor static quantization. For the per-token quantization, the combination with rotation-based methods can benefit in accuracy, but the randomness inherent in the input data prevents the prior calibration of quantization parameters, making it equally challenging to achieve the per-token static quantization.

In view of  the structured distribution of activation outliers, the per-channel quantization provides the possibility to study accurate 4-bit static quantization method. But, there are still some difficulties. In the scenarios involving per-channel quantization for both weights and activations, where the scaling factors $s^X\in \mathbb{R}^n$ and $s^W\in \mathbb{R}^j$  are employed, the computation for each element $Y$ of the linear layer output is expressed as follows:
\begin{equation}\label{c}
\begin{aligned}
Y_{ij} = s^W_j \left(\sum\nolimits_{k = 1}^n  s^X_k{{X_{ik}^{Int}}{W_{kj}^{Int}}}\right).
\end{aligned}
\end{equation}
Within this framework, the quantization scaling factors $s^X_k$ remain within the summation and cannot be extracted for external scaling in MM. This poses a challenge for the utilization of the integer acceleration kernels \cite{yuan2023rptq,wu2023understanding}, which typically require more simplified external scaling operations. In addition, Eq. \eqref{c} introduces an additional multiplication operation, resulting in approximately $\{n^3\}$ times extra multiplication operations over the entire matrix multiplication.

\section{Method}
\subsection{Quantization Step Migration}\label{4.1}
As pointed out in Sec. \ref{sec3}, calibrating the activation scaling factors along the channel dimension can better match the outlier distribution of LLMs, but does not align with the integer acceleration kernels. In this part, we propose the quantization step migration, which facilitates the computational overlap of the ``Quant/DeQuant" steps in the per-channel quantization with the Transformer block and match with integer acceleration kernels without changing the output.

\textbf{Quantization migration.} We calibrate the activation quantization scaling factors $s^{\tilde X}$ along the channel dimension using calibration samples from a small subset of the training dataset. For the computational overlap of ``Quant" steps, we find that the multiplier of RMSNorm provides an effective solution. Since the calibration of per-channel quantization occurs along the output dimension of RMSNorm, we can integrate the quantization scaling factors $s_k$ for the $n$-th channel dimension across various tokens with the multiplier parameters $\gamma_k$ corresponding to the output dimension of RMSNorm, as follows:
\begin{equation}\label{d}
\begin{aligned}
\tilde X_k^{Int}=\left\lceil\frac{\text{RMSNorm}(X_k)}{s_k^{\tilde X}}\right\rfloor = \left\lceil \frac{X_k}{\text{RMS}(X)}\frac{\gamma_k}{s_k^{\tilde X}} \right\rfloor.
\end{aligned}
\end{equation}
When dealing with the modification in LayerNorm, the principle of computational overlap in the ``Quant" steps still applies. By integrating the LayerNorm parameters (i.e., multiplier $\gamma_k$ and adder $\beta_k$) with the quantization scaling factors $s_k^{\tilde X}$ simultaneously, we can achieve computational overlap of the ``Quant" steps. This involves computing ${\gamma_k}/{s_k^{\tilde X}}$ and ${\beta_k}/{s_k^{\tilde X}}$ before the quantization.

\begin{figure*}[t]
    \centering
    \includegraphics[width=1.01\textwidth]{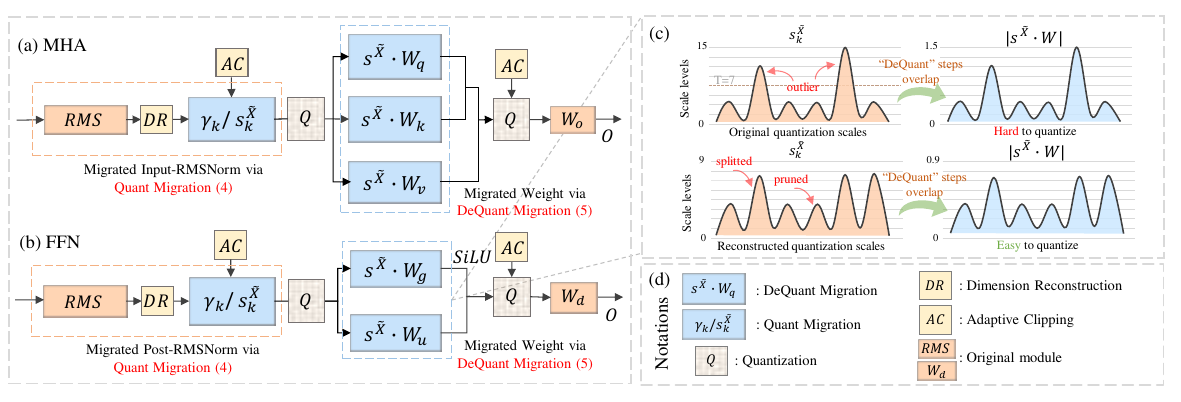}
    \caption{The overall framework of MergeQuant. (a,b): the integration of MergeQuant with a Llama layer, where quantization and dequantization migrations are represented in blue, dimension reconstruction and adaptive clipping are represented in yellow;  (c): The influence of ``DeQuant" step overlap on weight before and after dimension reconstruction; (d): necessary notations of MergeQuant.}
    \label{fig2}\vspace{-5pt}
\end{figure*}

\textbf{Dequantization migration.}
 For Eq. \eqref{c}, we cannot take the quantization scaling factors outside the matrix multiplication, which causes the ``DeQuant" steps of per-channel quantization to be repeated in the accumulator. We propose to overlap the computation of the ``DeQuant" steps by the linear mapping related to input activations, as follows:
\begin{equation}\label{e}
\begin{aligned}
Y_{ij} = s^{s^{\tilde X}\cdot W}_j \left(\sum\nolimits_{k = 1}^n  {{\tilde X_{ik}^{Int}}{(s^{\tilde X}_k \cdot W_{kj})^{Int}}}\right).
\end{aligned}
\end{equation}
We integrate the quantization scaling factors $s^{\tilde X}_k$, corresponding to the $k$-th activation channel,  with the weight parameters $W_k$ associated with the $k$-th input dimension. This integration enables us to seamlessly overlap the ``DeQuant" steps with the multiplication-addition operations in linear mapping,  thus avoiding explicit ``DeQuant" steps.

\textbf{Quantization mode under QSM.} Using the above migration pattern, we outline the per-channel quantization mode after QSM, as shown in Fig. \ref{fig2}. The activations are quantized as they pass through the RMSNorm multiplier. The RMSNorm outputs these activations in integer form after applying rounding. Subsequently, these integer activations are linearly mapped using a modified linear layer and are simultaneously dequantized. In fact, the calibration of quantization scaling factors and the integration of different modules can be performed before inference, so there will not increase the quantization computation overhead.

QSM provides an effective solution for the alignment of the per-channel static quantization with the integer acceleration kernels. The proposed quantization migration is near lossless, as the RSMNorm is always performed in FP16. However, the dequantization migration changes the  weight distributions along the input dimension, which adversely affects the weight quantization along the output dimension, as shown in Figure \ref{fig2}(c). 

\subsection{Quantization Scaling Modification}
In this part, we will introduce our dimension reconstruction and adaptive clipping methods, which mitigate the impact of dequant migration through simple dimension transformation and clipping without introducing additional computational overheads during inference.

\textbf{Dimension reconstruction.} Our goal is to mitigate the impact of dequantization migration on the weight quantization. A straightforward option is to increase the uniformity of static
 quantization scaling factors $s^{\tilde X}$, ensuring that the weights of different input dimensions have similar scaling. This can be achieved by splitting the activations or their static quantization parameters along the channel dimension. Note that the activations are dynamic and change with different inputs. Splitting the dynamic activations inevitably introduces extra computation and data movement overhead during inference. Therefore, we split the static quantization parameters and constrain them within a fixed range of $(0, T)$. Here we introduce a hyperparameter $\alpha$ to determine the split threshold $T$ based on the mean $\mu$ and variance $\sigma$ of the quantization parameters, using the following equation:  
\begin{equation}\label{f}
\begin{aligned}
Threshold: T = \mu(s^{\tilde X}) + \alpha\cdot \sigma(s^{\tilde X}).
\end{aligned}
\end{equation}
To limit the quantization scaling factors $s^{\tilde X}$ to the range of $(0, T)$, we select those scaling factors that exceed the threshold $T$ and designate them as ``strong parameters" $s_k$, which have the greatest impact on dequantization migration. After that, the strong parameter $s_k$ is decomposed into $(s-T,\ T)$. If $s_k^{\tilde X}-T$ is still greater than $T$, then decompose it into $(s_k^{\tilde X}-mT, \ T, \ \cdots)$ until $s_k^{\tilde X}-mT$ is less than $T$. We assume that the number of additional elements obtained after splitting all strong parameters is $M$.

After the splitting operation described above, the range of the static quantization parameter is limited to $(0, T)$, but this comes  at the cost of increasing its dimension to $n+M$. This irregular dimensional change is incompatible with the standard dimension required for efficient GPU computing kernel. Inspired by recent research \cite{guo2023olive} (i.e., the importance of normal channels adjacent to outlier channels is low), we propose a structured pruning method to restore the dimension of the quantization parameter. We identify the activation channels corresponding to strong parameters as outlier channels, and consider the following three cases to exclude repeated or non-existent neighboring channels: (1) outlier channels are adjacent; (2) there is a normal channel between two outlier channels; (3) the outlier channels are at the beginning or end of the sequence. Assuming that the number of neighboring channels obtained  under the above calculation is $N$. Combining the sensitivity of channel importance by Hessian matrix, we design the following three pruning schemes:

1. $N\!>\!M.$ Sort the diagonal elements of the Hessian matrix corresponding to the neighbor channels, and prune the quantization parameters corresponding to the first $M$ unimportant neighbor channels.

2. $N\!=\!M.$ Prune the quantization parameters corresponding to all neighbor channels.

3. $N\!<\!M.$ Sort the diagonal elements of Hessian matrix of channels other than the neighbor channels, and prune all neighbor channels and the first $M-N$  other channels.

\noindent\textbf{Adaptive clipping.}
In addition to the differences between activation channels, the large values that dominate the numerical range of the data space within each channel also have a significant impact on quantization scaling \cite{wei2022outlier,ashkboos2024quarot} and dequantization migration. Therefore, after the dimension reconstruction, we propose an adaptive clipping method that considers the per-channel pre- and post-quantization losses as well as the impact of dequantization migration on linear layers. When calibrating the per-channel static quantization parameters, we calculate the optimal clipping factor for each channel by measuring the difference between pre- and post-quantization activations and the quantization losses of transferred weights at different clipping ratios, given by:
\begin{equation}\label{f}
\begin{aligned}
L_i(s) = \Vert \hat{X}_i(s) - X_i(s)\Vert^2_F + \Vert \hat{W}^X - {W}^X\Vert^2_F,
\end{aligned}
\end{equation}
where $\hat{X}_i(s)$ represents the quantized channel activations, and $\hat{W}^X$ represents the weights of  after the dequantization migration.

For the down-linear layers in FNN and the out-linear layers in MHA, we do not observe obvious structured outliers. We use the per-token dynamic calibration and determine a uniform optimal clipping factor for all tokens by measuring the difference between the linear layer output before and after quantization under different clipping rates.

\noindent\textbf{ Implementation to Transformer blocks.}
The above methods have already limited the  range of quantization scaling factors. We then need to adjust the RMSNorm and linear layers to match the dimensions of the quantization parameters after dimension reconstruction. For the RMSNorm, we reconstruct the dimensions of the activations before the multiplier to ensure compatibility with the modified quantization parameters, as shown in fig \ref{fig2}. For the linear layer, we modify the weight matrix along the output dimension to accommodate the reconstructed activations. Specifically, we split the weight dimension corresponding to the strong parameters, and prune the weight dimension corresponding to the discarded quantization parameters. In Appendix \ref{SecC1}, we show the specific implementation of dimensional reconstruction and the significant advantages it brings compared with traditional cumbersome quantization operations.

\subsection{Quantization Compensation}
We apply LoRA quantization compensation parameter matrices $A \in \mathbb{R}^{M \times r}$ and $B \in \mathbb{R}^{r \times N}$ to compensate for the accuracy losses incurred during quantization and clipping. Specifically, the low-rank matrices $A$ and $B$ are embedded within each linear layer, and they are learned by minimizing the reconstruction error between the original output of the attention-FFN blocks and the quantized output. Ultimately, the quantized integer weight $q(W + AB)$ is constructed as the sum of the quantized weight and the compensation term  generated by the low-rank matrices.

\vspace{-5pt}
\renewcommand{\arraystretch}{1.15}
\begin{table*}[h!]
\caption{
The performance of MergeQuant framework is compared with the baselines on 5 different zero-shot tasks and 2 language datasets. MergeQuant$_{n-h}$ indicates that no additional hadamard rotations are introduced into model inference. } \label{tab1}
\centering
\scalebox{0.8}{{
\begin{tabular}{ccccc|c|ccccc|c}
\toprule
\textbf{Models} &\textbf{Method} & \textbf{Type}  & \textbf{WikiText-2} & \textbf{C4}& \textbf{Avg.$\downarrow$} & \textbf{PIQA} & \textbf{ARC-e} & \textbf{ARC-c} & \textbf{HellaSwag} & \textbf{Winogrande} & \textbf{Avg.(\%)$\uparrow$}\\
\midrule
\multirow{9}{*}{\shortstack{Llama\\2-7B}}
& FP16 & - & 5.47 & 7.26 &  6.22 & 79.05 & 74.54 & 46.16 &  75.98 & 69.06 & 68.95 \\
& SmoothQuant & static & 254.5 & - & - &  54.12 &  31.42 &  24.83 & 29.15 & 50.00 & 37.90 \\
& OmniQuant & dynamic & 14.61 & 18.39 & 16.50 & 65.94 & 43.94 & 30.80 & 53.53 & 55.09 & 49.86 \\
& QLLM  & dynamic & 11.75 & 13.26 & 12.51 & 67.68 & 44.40 & 30.89 & 58.45 & 56.59 & 51.60 \\
& QuaRot$_{n\!-\!h}$ & dynamic & 16.43 & - & - & 66.38 & 48.32 & 31.74 & 54.89 & 52.64 & 50.78 \\
& SpinQuant$_{n\!-\!h}$ & dynamic & 9.2 & - & - & 70.36 & 59.84 & 33.53 & 64.78 & 59.71 & 57.64 \\
& \cellcolor{myblue} MergeQuant$_{n\!-\!h}$ & \cellcolor{myblue} static & \cellcolor{myblue} 8.39 & \cellcolor{myblue}10.87 & \cellcolor{myblue}\textbf{9.63} &\cellcolor{myblue} 71.33 &\cellcolor{myblue}60.31 & \cellcolor{myblue} 34.22 &\cellcolor{myblue} 63.21 &\cellcolor{myblue} 58.88 &\cellcolor{myblue} \textbf{57.79}  \\
\cdashline{2-12}
& QuaRot & dynamic  & 6.10 & 8.32 & 7.21 & 76.33 & 68.35 & 42.32 & 72.53 & 65.11 & 64.92 \\
& SpinQuant & dynamic &  5.96 & 8.28 & 7.12 & 76.17 & 69.28 & 41.72 & 72.90 & 66.06 & 65.22 \\
&\cellcolor{myblue} {MergeQuant} & \cellcolor{myblue} static &\cellcolor{myblue} 6.09 & \cellcolor{myblue} {7.87} &\cellcolor{myblue} \textbf{6.98} &\cellcolor{myblue} 76.71 &\cellcolor{myblue}70.83 &\cellcolor{myblue} 44.11 &\cellcolor{myblue} 72.74 &\cellcolor{myblue} 66.61 &\cellcolor{myblue} \textbf{66.20}  \\
\midrule
\multirow{9}{*}{\shortstack{Llama\\2-13B}}
& FP16 & - & 4.88 & 6.47 & 5.68 & 80.47 & 77.44 & 49.15  & 79.39 & 72.14 & 71.71 \\
 & SmoothQuant & static & 254.53 & - & - &  62.47 &  45.22 &  27.13 & 44.33 & 50.78 & 37.90 \\
& OmniQuant & dynamic & 12.28 & 14.64 & 13.46 & 69.80 & 47.22 & 33.79 & 59.34 & 55.49 & 53.13 \\
& QLLM & dynamic & 9.09 & 11.13 & 10.11 &  70.46 & 48.48 & 34.39 & 62.80 & 55.41 & 54.31 \\
& QuaRot$_{n\!-\!h}$ & dynamic & 15.77 & - & - & 67.79 & 53.49 & 35.41 & 61.53 & 55.17 & 54.68\\
& SpinQuant$_{n\!-\!h}$ & dynamic & 7.20 & - & - & 73.25 & 67.32 & 38.21 & 70.61 & 65.21 & \textbf{62.92} \\
&\cellcolor{myblue} {MergeQuant$_{n\!-\!h}$} & \cellcolor{myblue} static & \cellcolor{myblue}{7.36} &\cellcolor{myblue}  9.50 & \cellcolor{myblue} \textbf{8.43} &\cellcolor{myblue} 73.34 &\cellcolor{myblue}63.55 &\cellcolor{myblue} 37.63 &\cellcolor{myblue} 66.51 &\cellcolor{myblue} 59.83 &\cellcolor{myblue} {60.17} \\
\cdashline{2-12}
& QuaRot & dynamic & 5.40 & 7.54 & 6.47 & 79.05 & 73.27 & 45.48 & 76.03  & 70.64 & 68.89 \\
& SpinQuant & dynamic &  5.24  & 7.48 & 6.36 & 78.56 & 75.97  & 46.76  & 77.01 & 67.88 & {69.24} \\
& \cellcolor{myblue} {MergeQuant} & \cellcolor{myblue} static & \cellcolor{myblue} 5.29 & \cellcolor{myblue} 6.98 & \cellcolor{myblue} \textbf{6.14} & \cellcolor{myblue} 78.62 & \cellcolor{myblue} 74.24 & \cellcolor{myblue} 47.27 & \cellcolor{myblue} 76.47 & \cellcolor{myblue} 70.01 & \cellcolor{myblue} \textbf{69.32} \\
\midrule
\multirow{9}{*}{\shortstack{Llama\\2-70B}}
& FP16 & - & 3.32 & 5.71 & 4.52 & 82.70 & 81.02 & 57.17 &  83.81 & 77.98 & 76.54 \\
 & SmoothQuant & static & 57.10 & - & - &   66.57 &  69.53 &  31.73 & 45.14 & 39.42 & 50.48 \\
& QLLM & dynamic  &  7.00 & 8.89 & 7.95 &  74.27 & 50.59 & 37.20 & 71.62 &  59.43 &  58.62 \\
& QuaRot$_{n\!-\!h}$ & dynamic  & 12.47 & - & - & 66.79 & 52.49 & 34.41 & 60.53 & 54.17 & 53.53\\ 
& \cellcolor{myblue} {MergeQuant$_{n\!-\!h}$} & \cellcolor{myblue} static & \cellcolor{myblue}{5.36} &\cellcolor{myblue}  7.50 & \cellcolor{myblue} \textbf{6.43} &\cellcolor{myblue} 76.34 &\cellcolor{myblue}71.55 &\cellcolor{myblue} 46.63 &\cellcolor{myblue} 73.51 &\cellcolor{myblue} 64.83 &\cellcolor{myblue} \textbf{66.58} \\
\cdashline{2-12}
& QuaRot & dynamic  &  3.79 & 6.12 & 4.96 & 81.83 & 79.76 &  55.46 &  81.58 & 76.09 & 74.94 \\
& SpinQuant & dynamic  & 3.70 &  6.07 & 4.89 & 81.61 & 79.17 & 56.31 &  82.36 & 75.85 & {75.06} \\
& \cellcolor{myblue}{MergeQuant} & \cellcolor{myblue} static  & \cellcolor{myblue} 3.78 & \cellcolor{myblue} 5.89 & \cellcolor{myblue} \textbf{4.84} & \cellcolor{myblue} 81.67 & \cellcolor{myblue} 79.89 & \cellcolor{myblue} 55.51 & \cellcolor{myblue} 81.91 & \cellcolor{myblue} 76.93 & \cellcolor{myblue} \textbf{75.19} \\
\midrule
\multirow{4}{*}{\shortstack{Llama\\3-8B}}
& FP16 & - & 6.14 & 9.45 & 7.80 & 80.74 &  77.57 & 53.50 & 79.12 &  72.93 & 72.77 \\
& QuaRot & dynamic  & 8.16 & 13.38 & 10.77 &  75.35 & 70.83 & 45.73 & 72.97 &  67.17 & 66.41 \\
& SpinQuant & dynamic  &  7.42  & 12.22 & 9.82 & 77.14 &  74.16 & 45.79 &  74.73 & 69.32 & {68.22} \\
& \cellcolor{myblue} {MergeQuant} & \cellcolor{myblue} static  &\cellcolor{myblue} 7.92 &\cellcolor{myblue} 11.71 &\cellcolor{myblue} \textbf{9.81} &\cellcolor{myblue} 77.96 &\cellcolor{myblue} 73.70 &\cellcolor{myblue} 46.38 &\cellcolor{myblue} 74.93 &\cellcolor{myblue} 69.58 &\cellcolor{myblue} \textbf{68.51} \\
\midrule
\multirow{4}{*}{\shortstack{Llama\\3-70B}}
& FP16 & - &  2.86 & 7.17& 5.02 &  84.44 & 85.94 & 64.25 &  84.93 & 80.74 & 80.06 \\
& QuaRot & dynamic  &  6.60 &  12.87 & 9.74 &   78.89 &  74.37 &  49.49 & 77.22 & 71.03 & 70.20 \\
& SpinQuant & dynamic  & 6.21 & 12.82 & 9.52 & 79.33 & 77.40 &  51.96 & 77.29 &  72.06 & \textbf{71.61} \\
&\cellcolor{myblue} {MergeQuant} & \cellcolor{myblue} static & \cellcolor{myblue} 6.86 &\cellcolor{myblue} 10.52  & \cellcolor{myblue} \textbf{8.69} &  \cellcolor{myblue} 79.82 &  \cellcolor{myblue} 73.11 &  \cellcolor{myblue} 50.00 &  \cellcolor{myblue} 77.71 &  \cellcolor{myblue} 72.35 &   \cellcolor{myblue} {70.59} \\
\bottomrule
\end{tabular}}}
\end{table*}

\section{Experiments}\label{sec4}
\textbf{Quantization settings.}
We use the per-channel static quantization for the activations with structured outliers and use GPTQ \cite{frantar2022gptq} as the standard method for per-channel weight quantization. For the activations without structured outliers, we apply per-token dynamic quantization to maintain model accuracy. All the quantization strategies are based on the max-min calibration method to calculate the maximum and minimum values of activations and weights to determine the appropriate quantization scaling factor. During the GPTQ quantization and LoRA \cite{hu2021lora, liu2023qllm} fine-tuning stages, we select 256 samples of the mixed datasets composed of WikiText-2 \cite{DBLP:journals/corr/MerityXBS16} and C4 \cite{10.5555/3455716.3455856} as the training sets.

\textbf{Models and datasets.}
We benchmark MergeQuant using Llama models \cite{touvron2023llama, dubey2024llama}, as they outperform other open-source models and serve as the foundation for many popular fine-tuning models. We select Wikitext-2  and C4 datasets to measure perplexity, which is closely related to performance on generation tasks. To further verify the accuracy of the quantized model across different tasks, we evaluate multiple zero-shot tasks, including PIQA \cite{articlePiqa}, HellaSwag \cite{zellers2019hellaswag}, WinoGrande \cite{10.1145/3474381}, ARC \cite{clark2018think}. These tasks  encompass various types such as common-sense reasoning, logical judgment, and reading comprehension. To ensure consistency and reliability of our experiment results, all evaluation experiments are conducted using lm-eval-harness \cite{eval-harness}. We use NVIDIA's CUTLASS library to accelerate Int4 matrix multiplication.

\textbf{Baselines.}
In our experiments, we first compare our method with the static quantization method SmoothQuant \cite{pmlr-v202-xiao23c}. We then compare the dynamic quantization methods, including OmniQuant \cite{shao2023omniquant} and QLLM \cite{liu2023qllm}, both of which employ learnable parameters and demonstrate strong performance on LLMs. For the rotation-based quantization methods, such as QuaRot \cite{shao2023omniquant} and SpinQuant \cite{liu2024spinquant}, the key idea is to reconstruct and suppress the structured outliers in activations, which are also included in our comparison. Except for the description in Table \ref{tab4}, all of these methods implement dynamic quantization by default. We provide two sets of comparisons based on whether to combine with additional hadamard rotation, as we find that it has a significant impact on different methods. 

\subsection{Accuracy Results}
\begin{figure*}[tbp]
    \centering
    \includegraphics[width=1\textwidth]{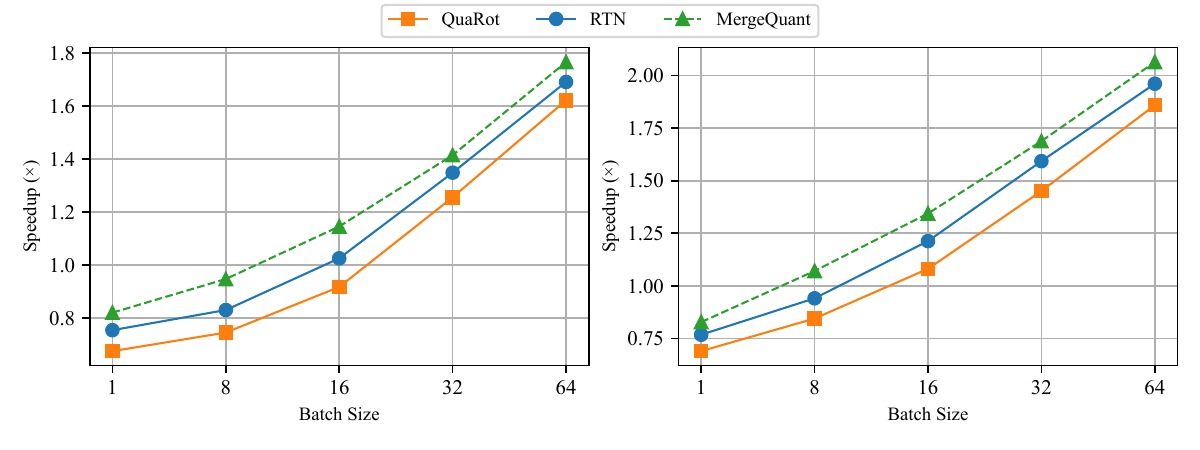}
    \begin{minipage}[t]{0.54\textwidth}
		\vspace{-18pt}        
		\centering 
        (a) Decoding speedup.
    \end{minipage}
    \hfill
    \begin{minipage}[t]{0.45\textwidth}
		\vspace{-18pt}  
        \centering
        (b) End-to-end speedup.
    \end{minipage}		\vspace{-15pt}
    \caption{For Llama-2-7B model, measure the decoding speedup and end-to-end speedup across various batch sizes. We pre-filled 2048 tokens and decoded 256 tokens. The experiments are performed on RTX 3090 GPUs.}\vspace{-10pt}
    \label{fig3}
\end{figure*}

\textbf{Results on language generation and zero-shot tasks.} 
We focus on evaluating the effectiveness of MergeQuant on Llama-2 and Llama-3 models using 4-bit weight and activation quantizations. Table \ref{tab1} gives the comparison between MergeQuant and baselines in terms of perplexity (language datasets) and accuracy (zero-shot tasks). We observe that MergeQuant is significantly closer to FP16 baseline than existing static quantization method across various Llama models. For example, on Llama-2 models, MergeQuant$_{n\!-\!h}$ achieves an accuracy improvement of approximately $20\%$ and reduces the perplexity by approximately $240$ compared to SmoothQuant. This verifies MergeQuant's effectiveness to maintain the model accuracy under 4-bit static quantization. For other dynamic quantization methods, we did not find the implementation of static quantization in the original papers, and we found in preliminary experiments that it would lead to a significant decrease in accuracy under the static quantization setting (see Table \ref{tab4}), so we do not provide their static quantization results.

We also compare MergeQuant with the dynamic quantization baselines, as shown in Table \ref{tab1}. We observe that these dynamic quantization methods gain from finer granularity calibration (per-token quantization) and dynamic calibration, achieving superior performance. However, our MergeQuant still outperform the competitive QuaRot and SpinQuant methods on Llama-2 models while also avoiding redundant quantization steps. Specifically, MergeQuant$_{n\!-\!h}$ reduces the average perplexity gap on WikiText and C4 datasets to 3.41 compared to FP16 implementation. When the additional Hadamard rotations are introduced, MergeQuant achieves an accuracy gap of only 2.4 points on Llama-2-13B model and an accuracy gap of only 1.3 points on Llama-2-70B model. Even on the more challenging-to-quantize Llama-3 models, MergeQuant can also achieve performance comparable to existing dynamic quantization methods. This demonstrates the effectiveness of our MergeQuant in balancing quantization accuracy and computational efficiency.

\renewcommand{\arraystretch}{1.02}
\begin{table}[t]\vspace{-5pt}
\caption{\label{tab2} Prefill  speedup of Llama-2-7B model across different batch sizes, compared to the baseline methods. The experiments were conducted on an NVIDIA RTX 3090 GPU. The sequence lengths set to 2048 tokens.}
\centering
\scalebox{1}{
\begin{tabular}{c|ccc} 
\toprule
\textbf{Batch Size} & \textbf{QuaRot} & \textbf{RTN}  & \textbf{MergeQuant} \\
\midrule
1  & 2.014x & 2.228x& 2.305x \\
8 & 2.123x & 2.402x  & 2.578x \\
16 & 2.157x & 2.444x  & 2.620x \\
32 & 2.170x & 2.470x  & 2.649x \\
64 & 2.209x & 2.515x & 2.715x \\
\bottomrule
\end{tabular}}\vspace{-5pt}
\end{table}

\begin{table}[t]
\caption{\label{tab3} Memory usage for decoding a single token with batch size of 1 and sequence length of 2048.}
\centering
\scalebox{1}{
\begin{tabular}{c|ccc}
\toprule
\textbf{Llama-2-7B}  & \textbf{QuaRot} & \textbf{RTN} & \textbf{MergeQuant} \\
\midrule
Memory usage & \multirow{2}{*}{\shortstack{4.16}}& \multirow{2}{*}{\shortstack{3.90}}  & \multirow{2}{*}{\shortstack{3.87}} \\ (GB) \\
Saving Factor  & 3.366x & 3.587x  & 3.583x\\
\bottomrule
\end{tabular}}\vspace{-5pt}
\end{table}

\subsection{Performance Analysis}\label{5.2}
\textbf{Decoding and end-to-end speedup.} 
This experiment aims to verify the efficiency of the proposed method in actual deployment on LLM. We pre-fill with 2048 tokens and decode 256 tokens. The primary baselines are simple RTN dynamic quantization and QuaRot dynamic quantization method. Figure \ref{fig3} shows that MergeQuant significantly outperforms  QuaRot in both decoding and end-to-end inference performance, and even surpasses the simple RTN dynamic implementation. Specifically, on Llama-2-7B model, our decoding acceleration reaches 1.765x and the end-to-end acceleration reaches 2.064x. This improvement is attributed to the proposed QSM method, which effectively avoids the need for explicit quantization step operations, with the only cost being a simple dimension reconstruction. Additional details are provided in Appendix \ref{SecC1}, where we visualize the runtime of both the dimension reconstruction and the dynamic quantization steps, highlighting the significant advantages of our method.

\renewcommand{\arraystretch}{1.15}
\begin{table*}[t]
\caption{\label{tab4} We analyze the impact of the proposed quantization step migration and adaptive clipping methods on different language datasets and zero-shot tasks under W4A4 quantization setting. The model used here is Llama-3-8B.}
\centering
\scalebox{1}{ 
\begin{tabular}{c|cc|ccccc} 
\toprule
\textbf{Methods} & \textbf{WikiText-2} & \textbf{C4}& \textbf{PIQA} & \textbf{ARC-e} & \textbf{ARC-c} & \textbf{HellaSwag} & \textbf{Winogrande} \\
\midrule
FP16 & 6.14 & 9.45 & 80.74 &  77.57 & 53.50 & 79.12 &  72.93 \\
\cdashline{1-8}
QuaRot \& Static & 18.23 & 23.56 & 62.02 &  47.05 & 28.67 & 50.03 &  57.22 \\
\cdashline{1-8}
+ QSM  & 10.61 & 15.42 & 72.03 & 61.41 & 37.88 & 67.47 &  65.67  \\
+ Clipping & 9.20 & 13.40 & 74.70 &  65.45 & 40.53 & 70.99 &  69.61  \\
+ Lora fine-tuning & 7.92 & 11.71 & 77.96 &  73.70 & 46.38 & 74.93 &  69.58 \\
\bottomrule
\end{tabular}}
\end{table*}

\textbf{Prefill speedup.} 
We investigate the actual acceleration performance of pre-filling with 2048 tokens, which is usually a computationally intensive task. We evaluate the acceleration performance of MergeQuant under different batch sizes, as detailed in Table \ref{tab2}. MergeQuant achieves a speedup of up to 2.715x, which is 0.506x higher than the QuaRot and 0.2x higher than the RTN dynamic implementation, which demonstrates the effectiveness of MergeQuant for handling computationally intensive tasks.

\textbf{Memory usage.}
In addition to evaluating the actual acceleration performance, we also assess the memory usage of MergeQuant on Llama-2-7B model, as shown in Table \ref{tab3}. Experimental results demonstrate that MergeQuant significantly reduces peak memory usage under W4A4 quantization settings, achieving a 3.58 times reduction compared to FP16 implementation. Compared with QuaRot, MergeQuant performs better in terms of memory saving, mainly due to avoiding the online dynamic quantization steps that usually increase the both computational and memory overhead. This reduction in memory consumption is of great significance for consumer-grade GPUs, because it leads to a more efficient inference and deployment of LLMs with limited hardware resources, thereby enhancing the overall performance and operational efficiency of the system.

\subsection{Ablation Studies}
\textbf{Illustrative analysis of MergeQuant.}
We investigate the impact of the proposed quantization step migration and adaptive clipping implemented in MergeQuant on quantization performance. Starting with W4A4 quantization of the Llama-3-8B model, we progressively apply different methods and evaluate their effectiveness on different language datasets and zero-shot tasks, as presented in Table \ref{tab4}. The results demonstrate that both the quantization step migration and adaptive clipping methods effectively reduce quantization losses. For the QuaRot method, the results of static per-tensor calibration results in considerable degradation of model performence. This is mainly due to the fact that MergeQuant avoids the dominance of outlier channels on quantization scale. By calibrating independent quantization factors to each channel, we can capture the differences among channels more accurately than per-tensor calibration. Additionally, incorporating LoRA fine-tuning leads to further performance improvements compared to no fine-tuning. In Appendix \ref{SecC2}, we provide a more detailed ablation study of the clipping component.

\textbf{Other quantization schemes.}
Here, we explore the impact of using 3-bit asymmetric and grouped quantization weights. Table \ref{tab5} shows that our method shows superior performance under different quantization strategies. This verifies the effectiveness of dimensionality reconstruction and adaptive cropping, and demonstrates the potential of the proposed method in low-bit weight quantization settings.

\begin{table}[h]\vspace{-10pt}
\caption{\label{tab5} For W3A4 quantization on Llama-2-7B model, results of perplexity and average accuracy results of five zero-shot tasks under different weight quantization settings.}
\centering
\scalebox{0.9}{
\begin{tabular}{c|ccc} 
\toprule
\textbf{Methods} & \textbf{WikiText-2} & \textbf{C4}  & \textbf{Avg.Acc.} \\
\midrule 
FP16 &  5.47 & 7.26 & 68.95\% \\
\cdashline{1-4} 
QuaRot-$w_{3-asym}$  & 7.16 & 9.34 & 61.24\% \\
QuaRot-$w_{3-group}$ & 7.04 & 9.17 & 62.08\% \\
\cdashline{1-4}
MergeQuant-$w_{3-asym}$ & 6.62 & 8.91  & 63.37\% \\
MergeQuant-$w_{3-group}$ & 6.37 & 8.51  & 64.58\% \\
\bottomrule 
\end{tabular}}
\end{table}

\section{Conclusion} 
This paper introduces MergeQuant, an innovative static quantization framework designed for LLMs, addressing the computational and memory challenges inherent in LLMs. MergeQuant focuses on the adaptability of per-channel quantization with the structured outliers of activations. By integrating the quantization step migration method, we significantly reduce the quantization overhead before and after matrix multiplication. Furthermore, we employ dimension reconstruction and adaptive clipping to mitigate the non-uniformity of the activation quantization scaling factors, while redistributing channel variations to the subsequent modules to balance the parameter distribution under QSM. Experimental results on different models show that MergeQuant can achieve up to 2.72× speedup in pre-filling, up to 1.77× speedup in decoding, and up to 2.06× speedup in end-to-end, while ensuring model accuracy. All performance evaluations are performed using a commercial RTX 3090 GPU, highlighting the practical applicability and accessibility of our approach in real-world applications.

\nocite{langley00}
\bibliography{custom}
\bibliographystyle{icml2025}

\newpage
\appendix
\onecolumn
\section{Related Work}
\noindent\textbf{Quantization.} 
Quantization has received extensive attention in the field of model compression as a key method to reduce the computational complexity and memory usage of deep neural networks (DNNs) \cite{nagel2020up}. By converting floating-point values to fixed-point or low-precision floating-point representations, quantization effectively reduces the storage requirements of model parameters and accelerates the model inference. Quantization can be mainly divided into two categories: Quantization-Aware Training (QAT) \cite{polino2018model,Jacob_2018_CVPR} and Post-Training Quantization (PTQ) \cite{bengio2013estimating,choi2018pact}.  QAT incorporates quantization constraints during training, which can improve the accuracy of the quantized model. However, it requires significant computing resources \cite{10.5555/3524938.3525605}, especially for LLMs that already have high training costs, making its application challenging. In contrast, PTQ quantizes pre-trained models without requiring additional training and is more suitable for the deployment of LLMs \cite{gholami2021survey}.

\noindent\textbf{Quantization of LLMs.} 
The substantial number of parameters in LLMs results in significant memory usage and computational overhead, posing challenges for their deployment and application. To address these challenges, quantization for LLMs have been extensively studied, primarily divided into two strategies: weight quantization \cite{wang2020towards,li2021brecq,sheng2023flexgen,dettmers2024qlora,lee2023owq,shen2020q} and weight-activation quantization \cite{luo2020positional,lin2020towards,choukroun2019low}.

Regarding weight quantization, the goal is to compress the model parameters to reduce memory usage while maintaining high precision of activations. GPTQ \cite{frantar2022gptq} employs a second-order approximation method to quantize the weights of LLMs to 4 bits, achieving low-bit quantization of the model without additional training. AWQ \cite{lin2023awq} introduces an activation-aware per-channel scaling strategy to further reduce quantization error during weight quantization. QUIP \cite{chee2024quip} utilizes random rotation matrices or Hadamard transforms to reduce quantization losses.

In terms of weight-activation quantization, the goal is to quantize both the weights and activations to achieve greater compression ratios and hardware acceleration. However, the structured outliers of activation in LLMs lead to a significant increase in quantization losses, which poses a significant challenge to the viability of low-bit activation  quantization \cite{kovaleva2021bert}.  LLM.int8() \cite{dettmers2022llmint8} adopts a mixed-precision approach to mitigate activation quantization errors by retaining outlier channels in activations at high precision (FP16). SmoothQuant \cite{pmlr-v202-xiao23c} proposes a per-channel scaling transformation to balance the numerical ranges between weights and activations, making the model more suitable for quantization. Outlier Suppression \cite{wei2022outlier} reduces the impact of activation outliers on quantization scaling by adjusting the scaling operation in LayerNorm. QLLM \cite{liu2023qllm} makes the distribution between channels more uniform by separating the outlier channels, thus reducing the dominance of outliers over quantization scaling. QuaRot \cite{ashkboos2024quarot} first applies Hadamard transform in W4A4 quantization of LLMs and proved its effectiveness in reducing quantization losses. SpinQuant \cite{liu2024spinquant} further introduces a learnable orthogonal matrix combined with a model-level loss function to mitigate the impact of outliers. However, these methods still face the challenge of maintaining model performance and hardware efficiency when implementing low-bit (e.g., 4-bit) static activation quantization.

\section{Implementation Details}
To collect the basic statistics required for static quantization and adaptive clipping, we perform a mixed analysis of the WikiText2 and C4 datasets, selecting 32 sentences of length 2048 from them as a calibration dataset. We conduct a simple search for the clipping ratio of each module and determine the maximum and minimum values of each channel based on the mean squared error (MSE). To achieve a reasonable balance between computational efficiency and model accuracy, we use the min-max calibration method for 4-bit weight quantization and all activation quantization within MergeQuant framework. For GPTQ quantization and LoRA fine-tuning, we randomly select 256 sentences of length 2048 from the WikiText2 and C4 mixed dataset as the training dataset, and adjust them over 15-epoch training process to compensate for the quantization errors and clipping losses. For  Llama-2 and Llama-3 models, we set the hyperparameters to $\alpha=5$ and $\alpha=2$, respectively.

Our scheme enables the unified application of static quantization and adaptive clipping across different models and ensures the pre-calibrating of quantization scaling factors and clipping factors. This design guarantees the versatility of the MergeQuant framework in various applications. With this unified quantization and clipping methods, we are able to fully evaluate the performance of the MergeQuant framework in different environments.

All model accuracy evaluations are performed using the PyTorch framework on 2 NVIDIA RTX A800 GPUs, ensuring the efficiency and reliability of the experiments. This configuration allows us to conduct large-scale experiments and conduct in-depth analysis of the actual effects of the framework.

\section{Supplementary Experiments}
\subsection{Performance Analysis: Detailed Results}\label{SecC1}

\begin{figure*}[t]
    \centering
    \includegraphics[width=1\textwidth]{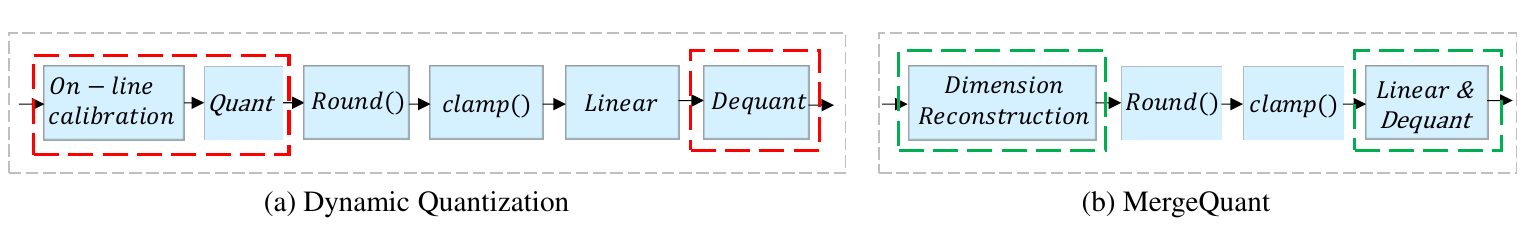}
		\captionsetup{skip=-5pt}
		\vspace{-0.8cm}
    \caption{The quantization process of Dynamic Quantization (a) and our MergeQuant (b).}
    \label{fig3.5}
\end{figure*}

Although simply converting weights and activations to INT4 using dynamic quantization methods can take advantage of integer acceleration kernels, this approach still introduces additional data movement overhead associated with quantization and dequantization operations, as shown in Fig. \ref{fig3.5} (red box). Such overhead is considerably expensive for autoregressive inference. To eliminate these costs, we use static calibration and QSM methods to avoid the quantization/dequantization steps required by dynamic quantization, at the cost of introducing only a simple dimension reconstruction operation, as shown in Fig. \ref{fig3.5} (green box).

For the dimension reconstruction of static quantization parameters and weights, we can carry out offline according to Sec 4.2. The dimension reconstruction of activations is introduced in model inference process, as follows:
\definecolor{lightgray}{gray}{0.95}
\begin{lstlisting}[basicstyle=\ttfamily\small, backgroundcolor=\color{lightgray}]
all_indices = torch.cat((keep_indices, position_tensor))
def Reconstructed_activation_matrix(activation, all_indices):
"""Extracts specific indices from the activation tensor along the last dimension."""
    original_shape = activation.shape
    Reconstruction_activation = activation[..., all_indices]
    if original_shape[:-1] != extended_activation.shape[:-1]:
       Reconstruction_activation = extended_activation.view(original_shape)
    return Reconstruction_activation
\end{lstlisting}
where "$position\_tensor$" represents the dimension index that needs to be extended, and "$keep\_indices$" represents the index that needs to be retained after excluding clip-on values, both of which can be obtained offline before inference.

Table \ref{tab6} presents the benchmarking results comparing our proposed dimension reconstruction to the dynamic quantization steps implemented in PyTorch. In our experiments, we selecte sequence lengths of 1, 128, and 256 to represent different stages of autoregressive inference: length 1 for the decoding stage and lengths 128 and 256 for the pre-filling stage, simulating the needs of real-world tasks. The experimental results show that, compared with the dynamic quantization method, our dimension reconstruction method significantly reduces latency across various sequence lengths and batch sizes. Specifically, we achieve up to 2.96x speedup over the dynamic quantization in decoding stage, and 2.41x speedup in pre-filling stage. All experimental evaluations are carried out on NVIDIA RTX 3090 GPU.

\begin{table*}[t]
\caption{\label{tab6}
Comparison of the dimension reconstruction method with the quantization strategy on NVIDIA RTX 3090 GPU, showing results for different batch sizes and different sequence lengths. We take the average of 500 runs and report numbers in milliseconds (ms).} 
\centering
\scalebox{1}{{
\begin{tabular}{c|c|c|cc|c}
\toprule
\textbf{Batch }&\textbf{\multirow{2}{*}{Hidden Sizes}} &\textbf{{Sequence}}  &  \textbf{{Dynamic}}& \textbf{Dimension} & \textbf{\multirow{2}{*}{Speedup}} \\
\textbf{Sizes} & &  \textbf{Lengths} & \textbf{{Quantization}} & \textbf{Reconstruction} &  \\
\midrule
\multirow{9}{*}{\shortstack{1}}&\multirow{4}{*}{\shortstack{4096}}
& 1& 0.070 & 0.024 &  2.92x \\
&& 128 & 0.070 & 0.029 & 2.41x \\
&& 256 & 0.083 & 0.043 & 1.93x  \\
\cdashline{2-6}
&\multirow{4}{*}{\shortstack{5120}}
& 1& 0.070 & 0.024 &  2.92x \\
&& 128 & 0.070 & 0.029 & 2.41x \\
&& 256 & 0.108 & 0.056 & 1.93x  \\
\cdashline{2-6}
&\multirow{4}{*}{\shortstack{8192}}
& 1& 0.070 & 0.024 &  2.92x \\
&& 128 & 0.082 & 0.045 & 1.82x \\
&& 256 & 0.142 & 0.087 & 1.63x  \\
\midrule
\multirow{9}{*}{\shortstack{16}}&\multirow{4}{*}{\shortstack{4096}}
& 1& 0.071 & 0.024 &  2.96x \\
&& 128 & 0.472 & 0.285 & 1.66x \\
&& 256 & 0.919 & 0.559 & 1.64x  \\
\cdashline{2-6}
&\multirow{4}{*}{\shortstack{5120}}
& 1& 0.070 & 0.024 &  2.92x \\
&& 128 & 0.579 & 0.372 & 1.56x \\
&& 256 & 1.139 & 0.730 & 1.56x  \\
\cdashline{2-6}
&\multirow{4}{*}{\shortstack{8192}}
& 1& 0.071 & 0.024 &  2.96x \\
&& 128 & 0.907 & 0.579 & 1.57x \\
&& 256 & 1.773 & 1.141 & 1.55x  \\
\midrule
\multirow{9}{*}{\shortstack{32}}&\multirow{4}{*}{\shortstack{4096}}
& 1& 0.071 & 0.024 &  2.96x \\
&& 128 & 0.920 & 0.560 & 1.64x \\
&& 256 & 1.785 & 1.109 & 1.61x  \\
\cdashline{2-6}
&\multirow{4}{*}{\shortstack{5120}}
& 1& 0.071 & 0.024 &  2.96x \\
&& 128 & 1.139 & 0.730 & 1.56x \\
&& 256 & 2.221 & 1.441 & 1.54x  \\
\cdashline{2-6}
&\multirow{4}{*}{\shortstack{8192}}
& 1& 0.071 & 0.024 &  2.96x \\
&& 128 & 1.773 & 1.146 & 1.55x \\
&& 256 & 3.507 & 2.271 & 1.54x  \\
\bottomrule
\end{tabular}}}
\end{table*}

\subsection{Ablation study of MergeQuant’s clipping component}\label{SecC2}
Table \ref{tab7} provides the perplexity results under different clipping strategies when only 4-bit activation quantization is applied. The results indicate that applying clipping before calibration can effectively reduce quantization losses without increasing computational burden, and this effect is particularly evident on Llama-3 models. For example, on Llama-2-7 model, clipping the qkv, up, and gate linear layers before per-channel calibration reduces the perplexity by about 0.4, whereas on Llama-3-8B model, the reduction is about 0.9. In addition, clipping the out and down linear layers can further reduce the difference between the quantized model and  FP16 representation.

\begin{table*}[t]
\caption{\label{tab7}
Ablation study of MergeQuant’s clipping component on Llama models.} 
\centering
\scalebox{1}{{
\begin{tabular}{c|c|cc|c}
\toprule
\textbf{Models} &\textbf{Method}  & \textbf{WikiText-2} & \textbf{C4}& \textbf{PPL-Avg.$\downarrow$} \\
\midrule
\multirow{4}{*}{\shortstack{Llama-2-7B}}
& FP16& 5.47 & 7.26 &  6.22 \\
& No-clipping & 7.32 & 8.96 & 8.14 \\
& Channel-clipping & 6.91 & 8.58 & 7.75  \\
& Adaptive clipping & 6.46 &  8.19  & 7.33\\
\midrule
\multirow{4}{*}{\shortstack{Llama-2-13B}}
& FP16 & 4.88 & 6.47 & 5.68 \\
& No-clipping & 5.78 & 7.48 & 6.63  \\
& Channel-clipping & 5.35 & 7.03 & 6.19 \\
& Adaptive clipping & 5.30 & 6.98 & 6.14 \\
\midrule
\multirow{4}{*}{\shortstack{Llama-2-70B}}
& FP16 & 3.32 & 5.71 & 4.52 \\
& No-clipping & 4.23 & 6.64 &  5.43 \\
& Channel-clipping & 3.96 & 6.14 & 5.05 \\
& Adaptive clipping & 3.89 & 6.06 & 4.97 \\
\midrule
\multirow{4}{*}{\shortstack{Llama-3-8B}}
& FP16 & 6.14 & 9.45 & 7.80 \\
& No-clipping & 9.64 & 13.45 & 11.55 \\
& Channel-clipping & 8.77 & 12.68 & 10.72 \\
& Adaptive clipping & 7.96 & 11.61 & 9.79 \\
\midrule
\multirow{4}{*}{\shortstack{Llama-3-70B}}
& FP16  &  2.86 & 7.17& 5.02  \\
& No-clipping & 11.34 & 14.03 & 12.68  \\
& Channel-clipping & 9.01 & 11.97 & 10.49   \\
& Adaptive clipping & 7.12 & 10.73 & 8.93  \\
\bottomrule
\end{tabular}}}
\end{table*}

\subsection{Quantization Time}
MergeQuant eliminates the need for full retraining and requires only brief parameter fine-tuning to maintain model performance. Compared to methods that require full parameter retraining, our framework is significantly more efficient in terms of computational time and resource utilization. As shown in Table \ref{tab8}, MergeQuant efficiently calculates the scaling factors and static quantization parameters for the Llama-3-70B model in only 44.5 minutes. Additionally, it takes only 9.3 hours to complete the overall quantization of the model parameters. This fast processing capability makes MergeQuant a powerful solution for deploying LLMs in resource-constrained environments.

\begin{table}[t]
\caption{\label{tab8}
 MergeQuant runtime for full quantization of Llama models.}
\centering
\scalebox{0.9}{
\begin{tabular}{llllllllllllllllll}
\hline
 \multicolumn{1}{c}{{Models}}  & \makecell[c]{{Llama-2-7B}} & \makecell[c]{{Llama-2-13B}} & \makecell[c]{{Llama-2-70B}} & \makecell[c]{{Llama-3-8B}} & \makecell[c]{{Llama-3-70B}}  \\
\hline
 \makecell[c]{Calibration} & \makecell[c]{{5.3m}} & \makecell[c]{{9.2m}} &\makecell[c]{{43.1m}} &  \makecell[c]{{5.9m}} &  \makecell[c]{{44.5m}} \\

 \makecell[c]{Fine-Tuning} & \makecell[c]{{1.6h}} & \makecell[c]{{2.8h}} &\makecell[c]{{9.3h}} &  \makecell[c]{{1.7h}}   &  \makecell[c]{{9.3h}}\\
\hline
\end{tabular}}
\end{table}

\section{More Visualizations} \label{SecD}
\begin{figure*}[h]
    \centering
    \includegraphics[width=1\textwidth]{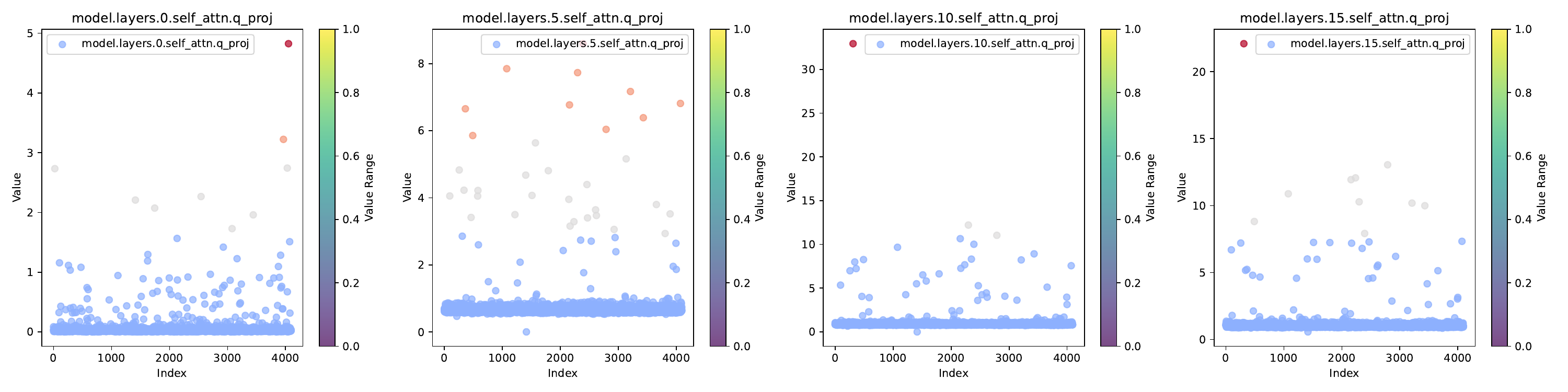}
    \includegraphics[width=1\textwidth]{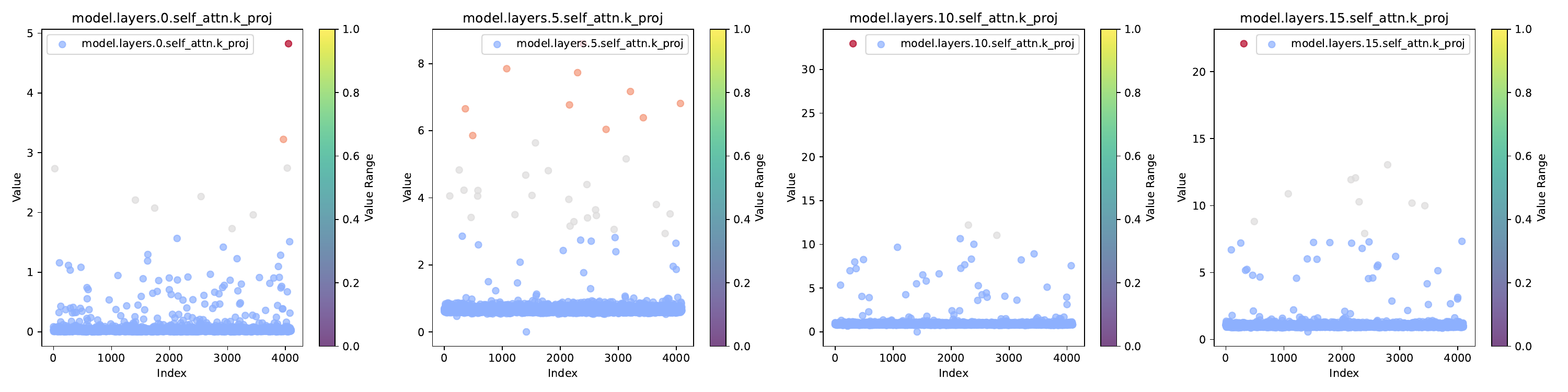}
    \includegraphics[width=1\textwidth]{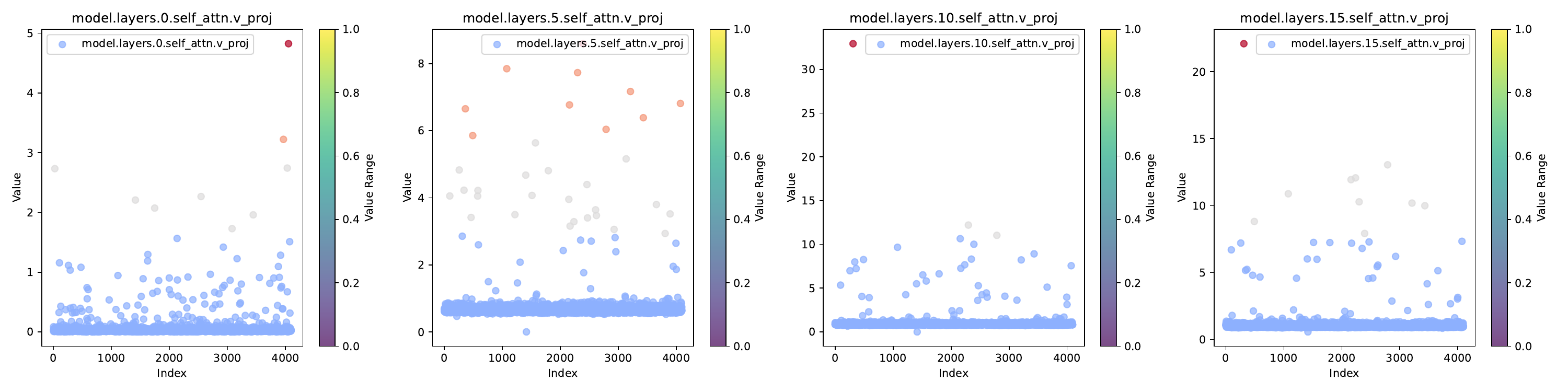}
    \includegraphics[width=1\textwidth]{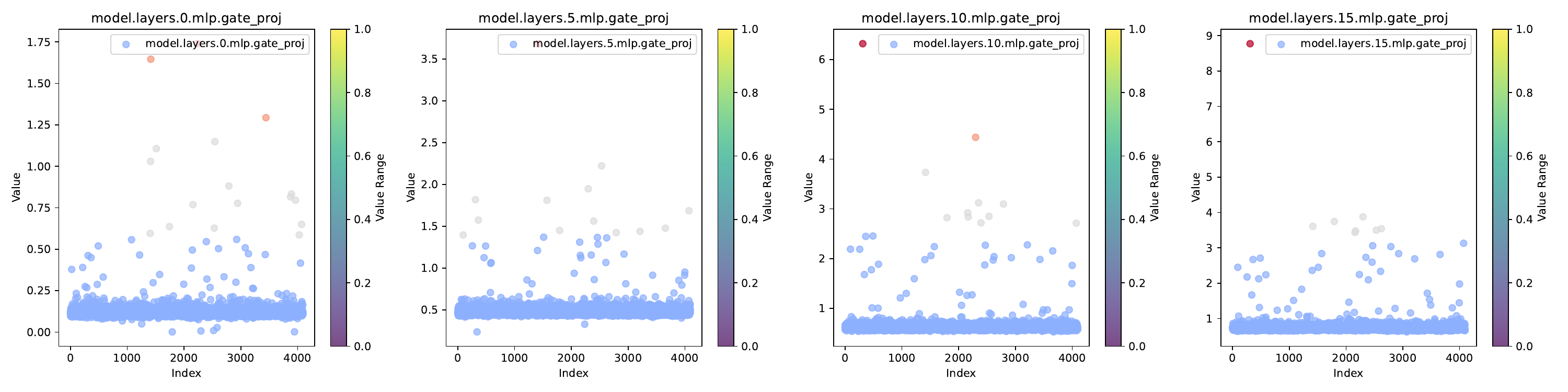}
    \includegraphics[width=1\textwidth]{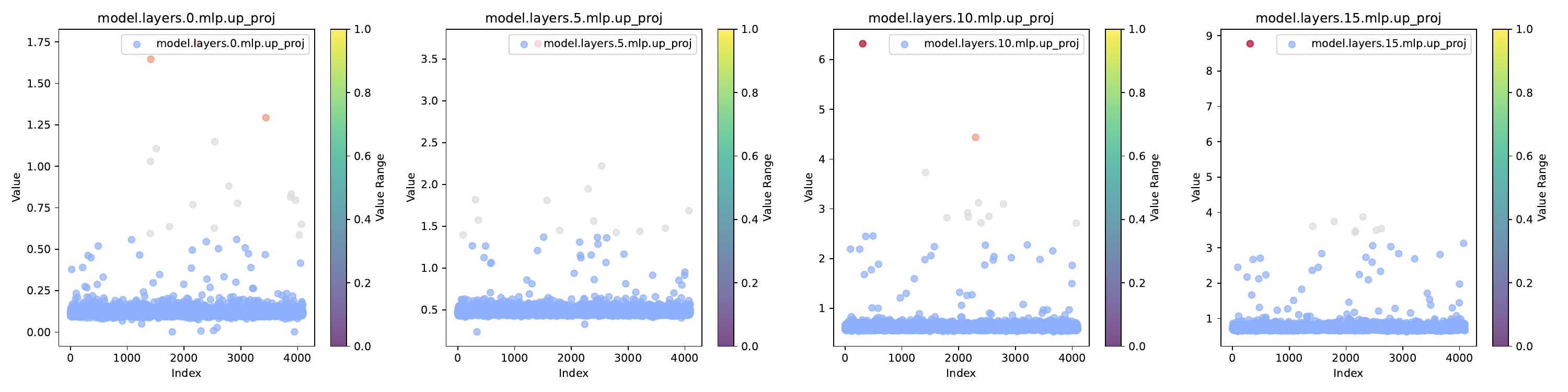}
    \caption{Visualization of Maximum absolute values in linear layers of Llama-2-7B}
    \label{fig4}
\end{figure*}

\begin{figure*}[h]
    \centering
    \includegraphics[width=1\textwidth]{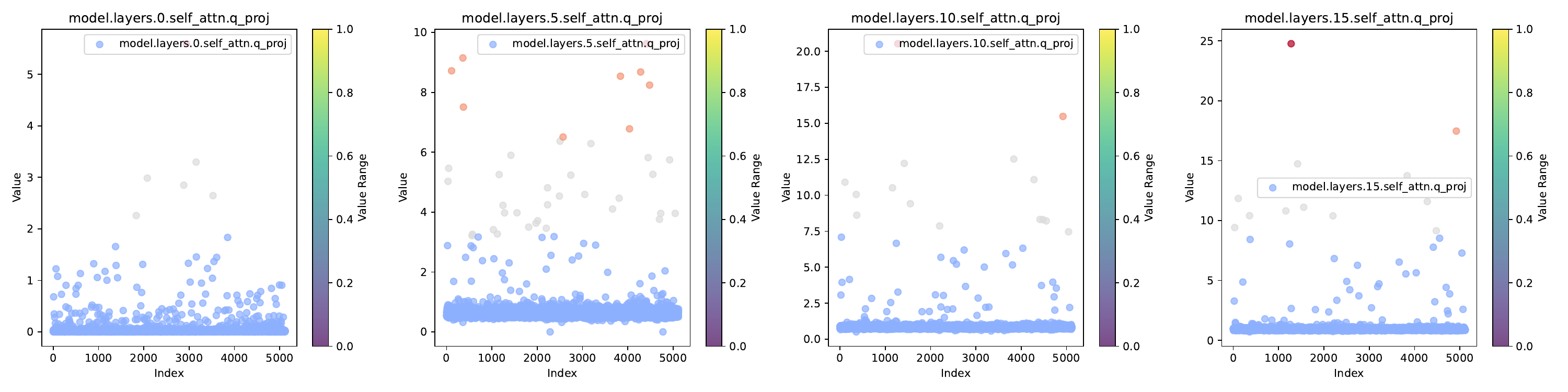}
    \includegraphics[width=1\textwidth]{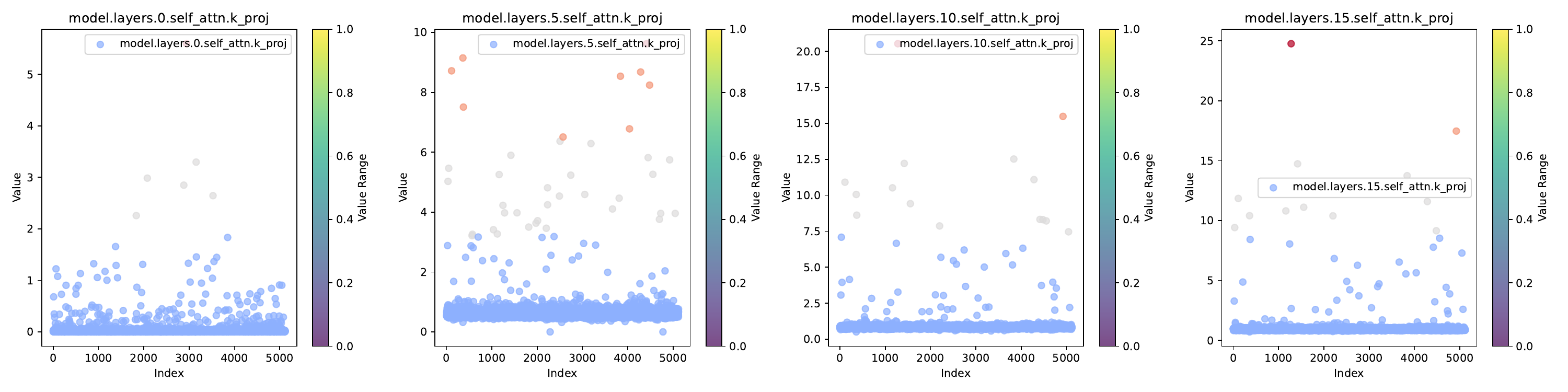}
    \includegraphics[width=1\textwidth]{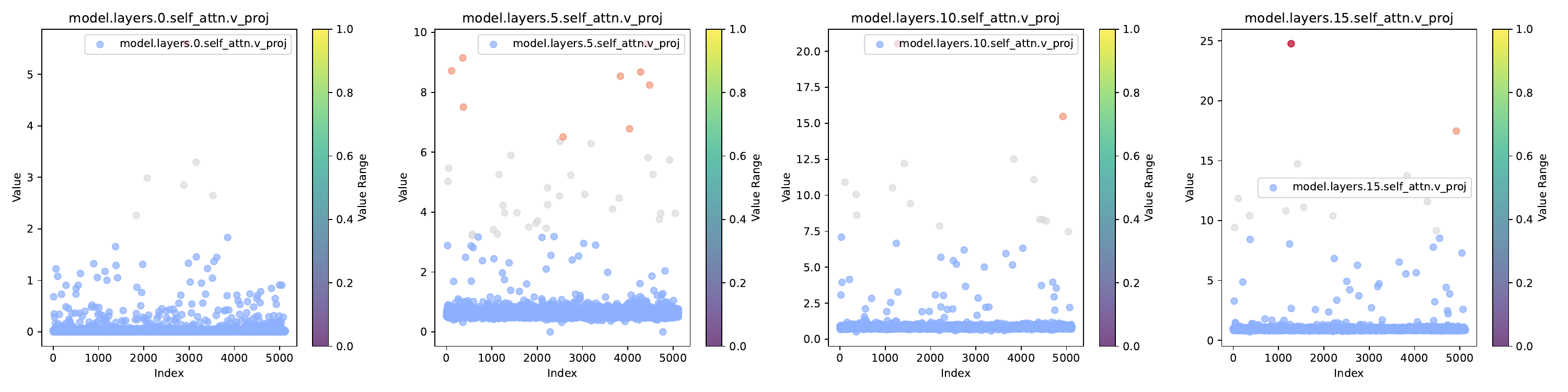}
    \includegraphics[width=1\textwidth]{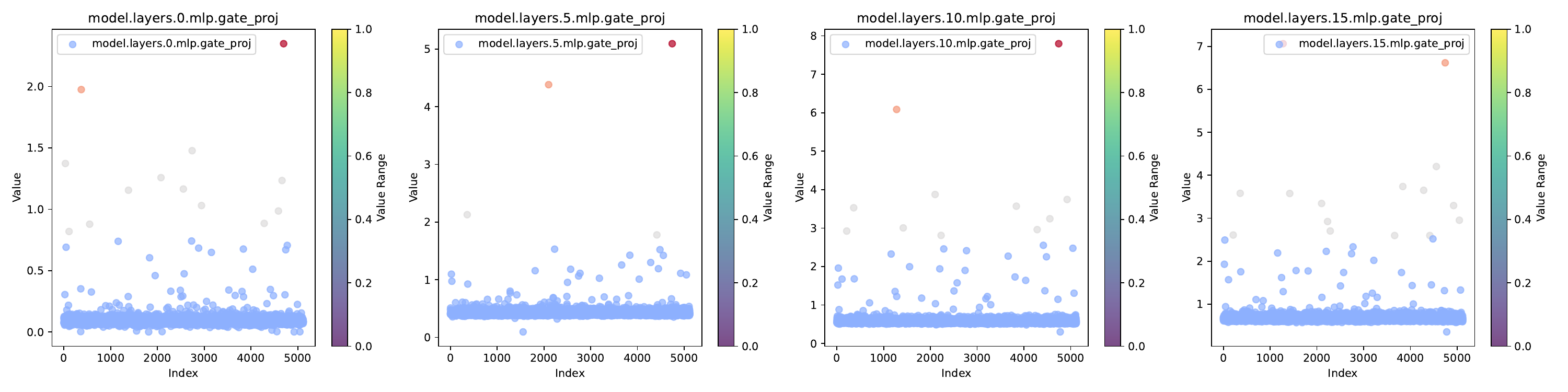}
    \includegraphics[width=1\textwidth]{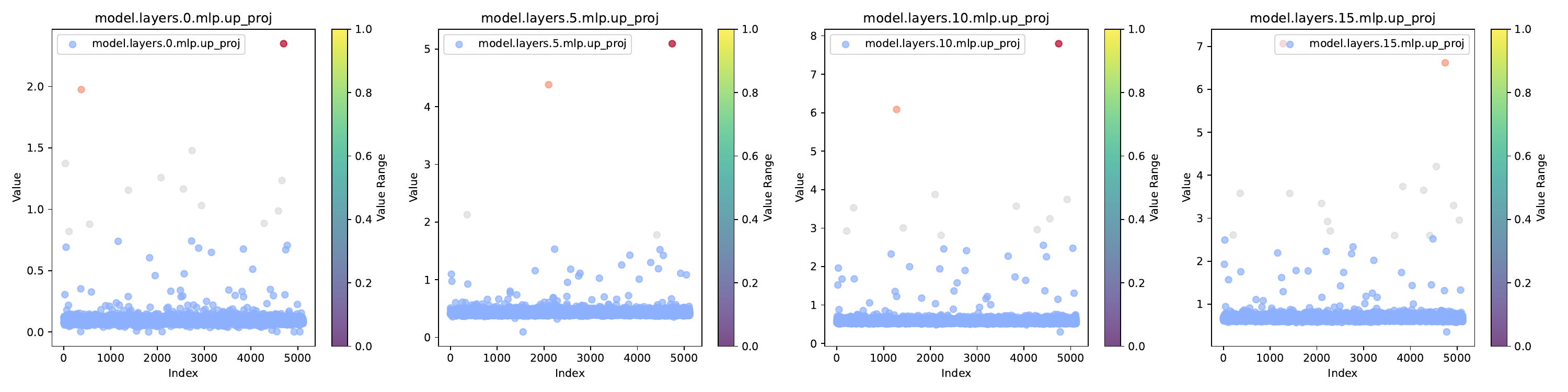} 
    \caption{Visualization of Maximum absolute values in linear layers of Llama-2-13B}
    \label{fig5}
\end{figure*}

As shown in Fig.\ref{fig4} and Fig.\ref{fig5}, we visualize the maximum absolute values of all dimensions in qkv, up, and gate linear layers of different Llama models, revealing the structured outlier patterns consistent with previous findings. That is, most of the outliers in the activation only exist in a small number of fixed channels. Our proposed per-channel calibration method effectively avoids interference between different channels in the quantization, thereby avoiding the dominance of outlier channels on quantization scaling.

\begin{figure}[h]
  \centering
  \begin{minipage}[t]{1\textwidth}
    \centering
    \includegraphics[width=15cm]{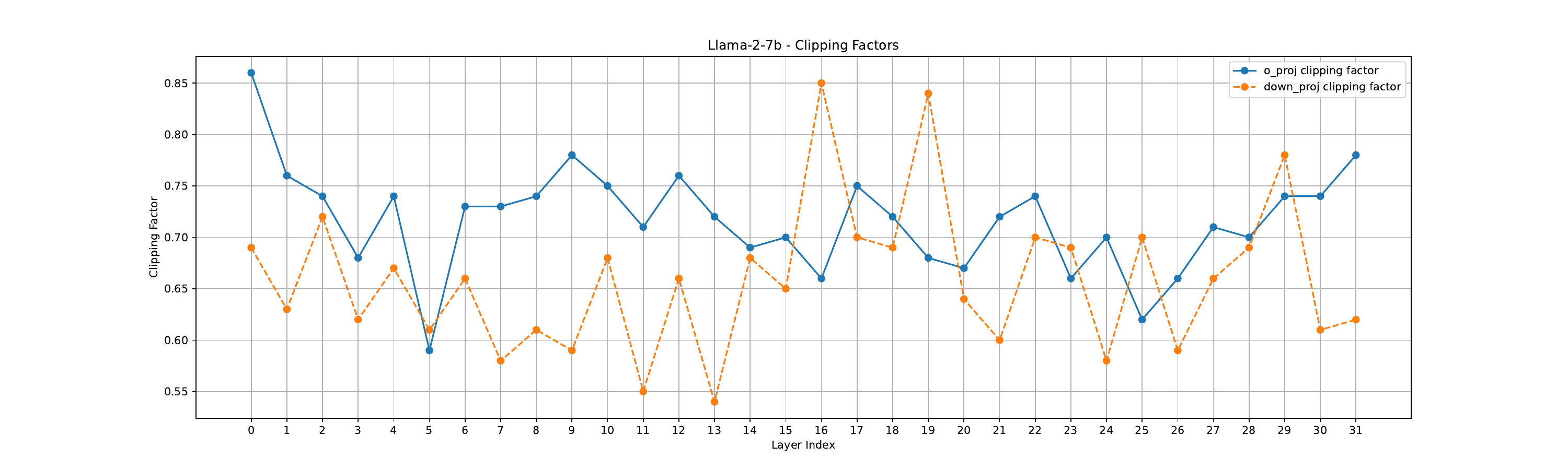}
  \end{minipage}\vspace{-2mm} 
  \begin{minipage}[t]{1\textwidth}
    \centering
    \includegraphics[width=15cm]{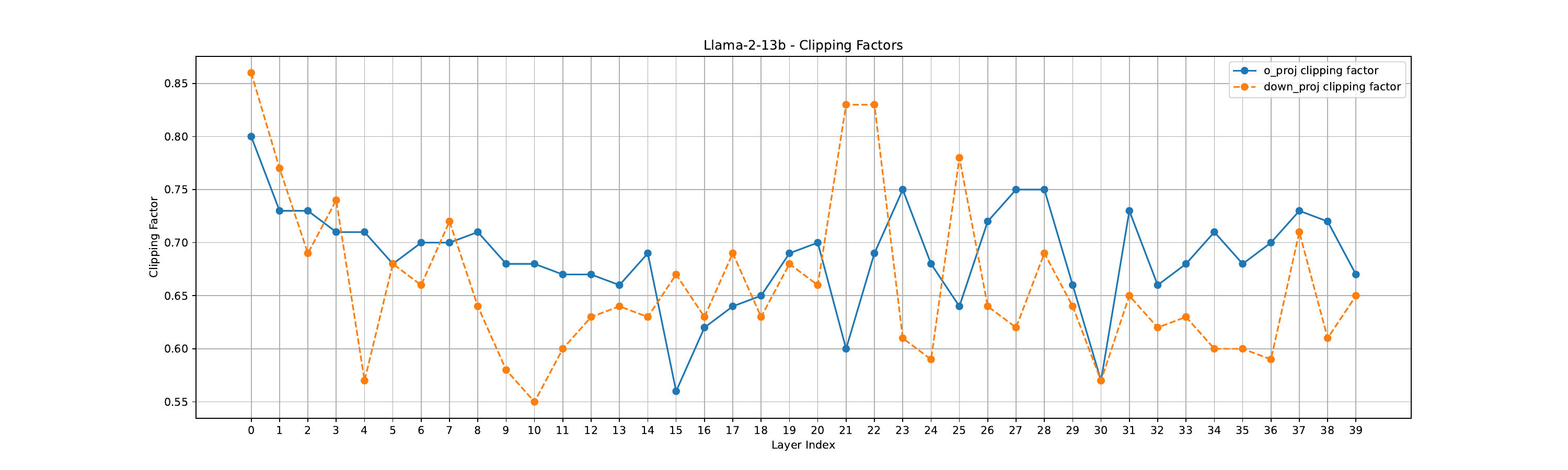}
  \end{minipage}\vspace{-2mm} 
  \begin{minipage}[t]{1\textwidth}
    \centering
    \includegraphics[width=15cm]{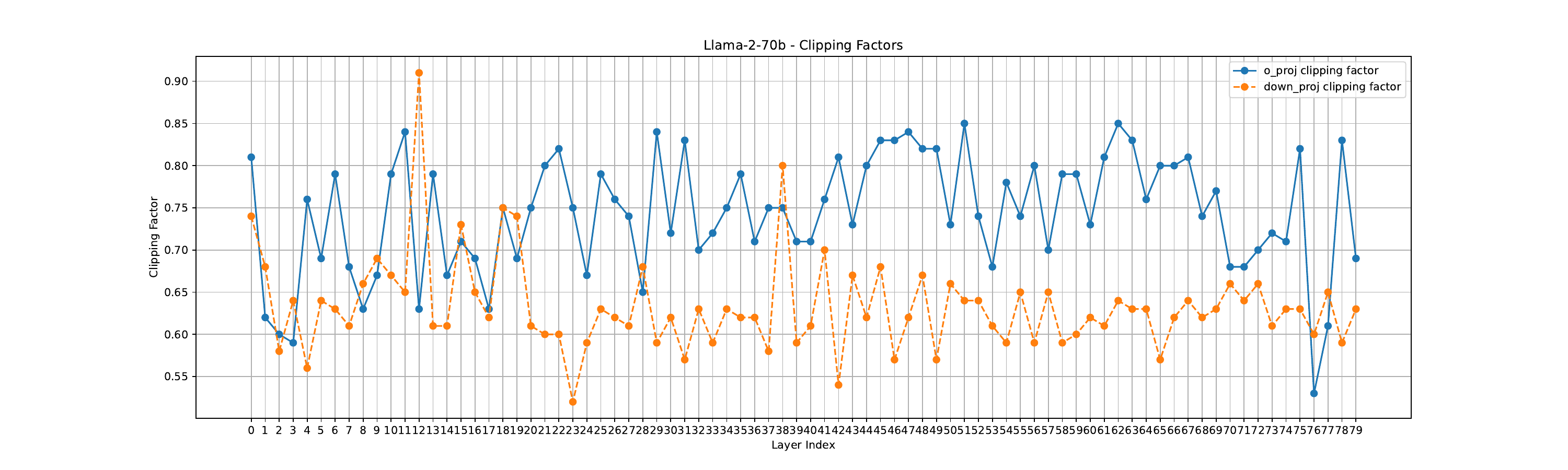}
  \end{minipage}\vspace{-2mm} 
  \centering
  \begin{minipage}[t]{1\textwidth}
    \centering
    \includegraphics[width=15cm]{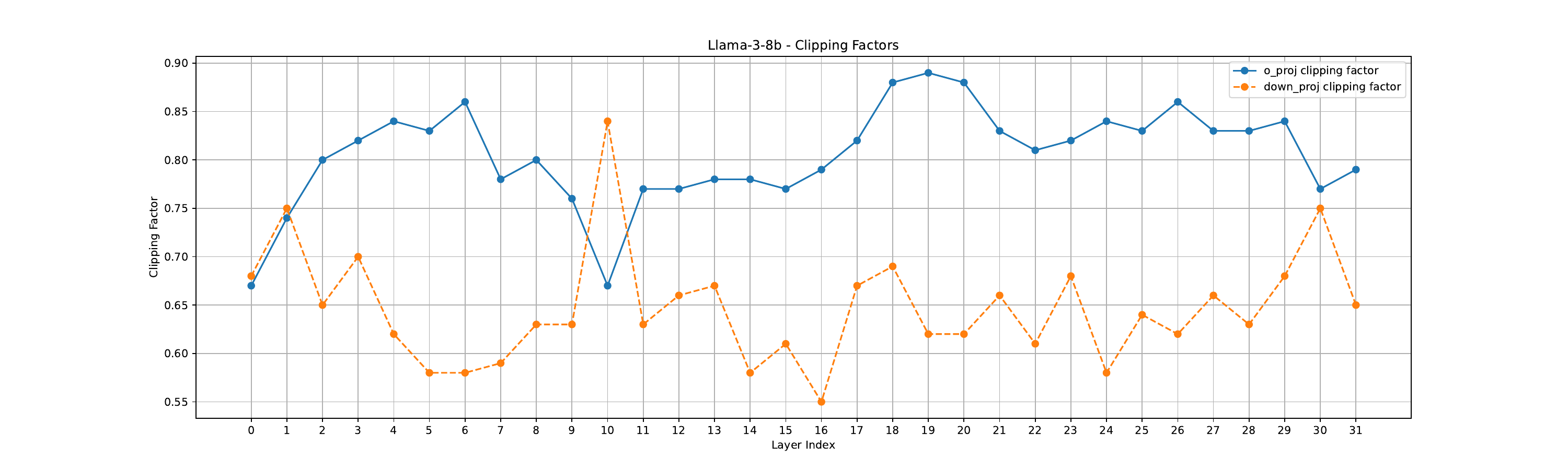}
  \end{minipage}\vspace{-2mm} 
  \begin{minipage}[t]{1\textwidth}
    \centering
    \includegraphics[width=15cm]{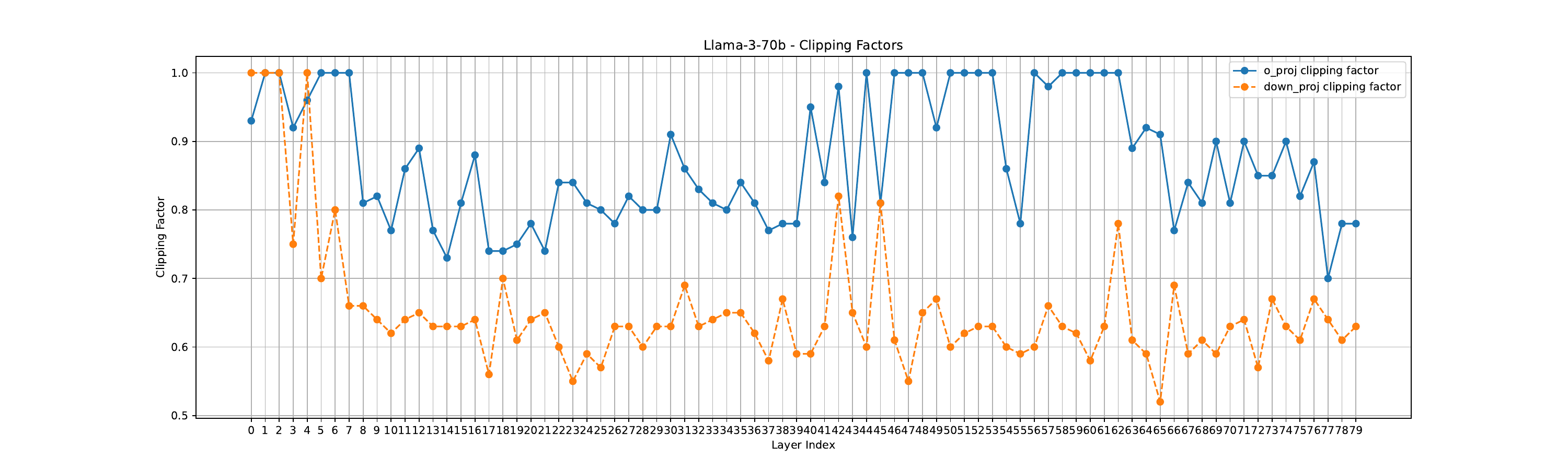}
  \end{minipage}
\caption{\label{fig6}
Activation clipping ratios for out and down linear layers in the Llama family. }
\end{figure}

Figure \ref{fig6} illustrates the activation clipping ratios of the out and down linear layers on Llama models. For the out linear layers, most clipping ratios are set between 0.7 and 0.8, while for the down linear layers, more aggressive clipping is applied, with most ratios set between 0.6 and 0.7.


\end{document}